%% file: main.tex
\documentclass{article} 
\usepackage{iclr2027_conference,times}

\input{math_commands.tex}

\usepackage{hyperref}
\usepackage{url}
\usepackage{booktabs}
\usepackage{graphicx}
\usepackage{caption}
\usepackage{multirow}

\title{Your Prompt Should Do More: Effects of \mbox{Retrieval} Instructions in Embedding Models}

\author{Amanda Myntti*, Jenna Kanerva*, Veronika Laippala, Filip Ginter$^\dagger $
\\
TurkuNLP\\
University of Turku, Finland\\
*Equal contribution \\
$^\dagger$Ellis Institute Finland \\
\texttt{\{amanda.a.myntti,jmnybl,mavela,figint\}@utu.fi} \\
}

\iclrfinalcopy 
\begin{document}

\maketitle

\begin{abstract}

Prompted embedding models have recently received increasing attention, particularly for retrieval, where detailed retrieval instructions are provided as part of the retrieval prompt. Several new datasets and studies have examined this setting, showing that the current embedding models often struggle to follow such instructions reliably.
In this paper, we study the mechanism of how instructions actually affect the representations of retrieval queries in asymmetric retrieval tasks. We show that models can fail to follow even simple task instructions when query-side distractors are included in the evaluation. We hypothesize that this behavior is driven by the training setup of current embedding models and their evaluation, and show that fine-tuning with added query-side distractors leads to substantial improvements, with minimal effect on other tasks. 
\end{abstract}

\section{Introduction}

Text embeddings are widely used for semantic matching, where relevance of two texts is determined by their embedding similarity. Modern embedding models are often trained to support multiple tasks through natural language instructions that specify the intended relation between a query and its target. Typically, retrieval tasks are asymmetric, where only the query receives a task instruction while candidate documents are embedded independently. This allows a shared document index for multiple retrieval tasks. Yet, standard evaluation does not match this use. Popular benchmarks, like MTEB \citep{muennighoff-etal-2023-mteb}, measure performance on multiple embedding task categories, which suggests that models should adapt query representations toward task-specific targets. However, the tasks are evaluated in isolation: both the queries and candidate sets are constructed for a single, specific task. Consequently, candidate pools rarely contain semantically related texts that do not satisfy the task instruction. We argue that this setting provides limited insight into how instructions actually transform query embeddings and alter retrieval behavior.


In this paper, we study how instructions affect retrieval behavior and embedding representations, uncovering several unintuitive results. For example, task instructions do not always increase the similarity between a question and its answer, and irrelevant task prompts are not consistently distinguishable from relevant ones in terms of performance, suggesting that instructions only weakly affect query representations. We hypothesize that this behavior may reduce robustness when the candidate pool includes examples semantically similar to the queries, which may be the case in practical large-scale retrieval applications. Using a controlled evaluation setting, we show that semantic similarity can dominate retrieval irrespective of the relation required by the task at hand, and that models fail to follow instructions adequately. We find evidence that this behavior may be related to common contrastive training practices that lack query-side negatives, and show that fine-tuning with such negatives substantially improves robustness.

\section{Related Work}

\textbf{Text embedding models} Modern models rely on bi-encoder architectures trained on contrastive objectives \citep{reimers-2019-sentence-bert-sentence-transformers,gao-etal-2021-simcse}, a paradigm that continues to produce state-of-the-art embedding models to this day \citep{qwen3embedding}. In-batch negatives are often supplemented with hard negatives mined from high-scoring corpus documents \citep{xiong2021approximate}, sometimes with similarity-margin filtering to avoid false negatives \citep{moreira2024nvretrieverimprovingtextembedding, lee2025nvembed}. Recently, LLMs have also been used to generate synthetic hard negatives \citep{li2024synegllmdrivensynthetichardnegatives}. Another fundamental development is the introduction of \textit{asymmetric natural language instructions}, where a task-specific instruction is prepended to the query but not the target \citep{wang2024multilingual-e5}, which allows one model to adapt to multiple retrieval tasks. Following earlier work on instruction and prompt-based learning \citep{Brown-2020-LM-are-fewshot-learners,jiang-2022-promptbert,shin-2020-autoprompt}, embedding models have increasingly incorporated task instructions and synthetic data to improve instruction following \citep{dai-2023-promptagator,wang-etal-2024-improving-text}.

\textbf{Instructed Retrieval} Several datasets have been introduced to study instruction-following IR models \citep{oh2024instructir,weller-etal-2025-followir,weller2025mfollowir,zhou2025beyond,song-etal-2025-ifir,weller2025promptriever,zhuang2025betterinstructionfollowingretrieval}, showing that models often struggle to follow the instructions. Instructions in these datasets cover various aspects, ranging from preferred source types to expert healthcare criteria, but are often complex and query-specific, with a separate instruction for each query. Therefore, they are not designed to measure systematic task-level effects, making it difficult to analyze changes in the representations. In a similar vein, \citet{asai-etal-2023-berri} introduce an annotation scheme for targets which allows the instruction to define which documents are relevant.

\textbf{Study of representations} The effects that the inputs have on representations have been studied by \citet{yamagiwa-etal-2023-discovering}, who find interpretable axes in embeddings, \citet{lee2025shared}, who study geometry in LM input and output embeddings, \citet{chang-etal-2022-geometry}, who study how language is encoded in embeddings, and \citet{lin-etal-2024-towards-understanding}, who study the relation of input and resulting representation in jailbreak settings. \citet{gonzalezgutierrez-2025-promptsreshaperepresentationsempirical} recently show that unrelated instructions can perform well on embedding evaluation tasks, and that while instructions cause shifts in representations, the causes, mechanisms, and the structure of these changes remain unknown.

\textbf{Our contribution} Benchmark evaluations indicate which models perform the best, but offer little insight into \textit{why}. Related work demonstrates that instructions have complex, yet little studied effects on embedding representations. We contribute by analyzing the effects of roughly 50 diverse instructions with structural measures that quantify the shifts which the instructions introduce to query embeddings. Additionally, we identify a core weakness in the standard fine-tuning and evaluation setups: the lack of \textit{query-side negatives}. Current frameworks select negatives from the \textit{corpus} side and evaluate tasks in isolation, which we hypothesize leads to models that react to instructions inadequately. We validate this hypothesis by introducing \textit{distractors}, query-side hard negatives, both in fine-tuning and evaluation, showing that training with distractors leads to substantially stronger and more consistent instruction-following behavior.

\section{Methods}

Our experiments consist of the following steps: first, we analyze roughly 50 natural language instructions, examining how task-relatedness and structural measures relate to their performance. We then identify differences between bitext-mining and retrieval tasks, and introduce distractors (query-side negatives) to show that instructions generally do not affect the query embeddings adequately. Next, we fine-tune models using distractors to improve instruction-following while preserving performance on other tasks, and finally, we analyze the structural effects fine-tuning had on the models.

\subsection{Datasets, Models, and Instructions}

\textbf{Datasets} \ \ The datasets in this study are Tatoeba \citep{tatoeba} and ARCChallenge \citep{clark2018ARC, xiao2024ReasoningAsRetrievalBenchmark}, as well as SQuAD.v1 \citep{rajpurkar-etal-2016-squad}, which is used for the fine-tuning experiment. Tatoeba and ARCChallenge are in the MMTEB benchmark \citep{muennighoff-etal-2023-mteb, enevoldsen-2025-mmteb}: The Tatoeba dataset comprises translation pairs for a bitext-mining task (from English to target language), and ARCChallenge consists of scientific questions and their multiple-choice answers for a retrieval task. To mitigate the possibility of data contamination affecting our results, we omit popular retrieval datasets such as Natural Questions \citep{kwiatkowski-2019-natural-questions} and FEVER \citep{thorne-2018-fever}, on which the models we test have been reportedly fine-tuned. We use recall@1 to evaluate Tatoeba and ndcg@10 for ARC\-Challenge, matching the metric used by MTEB.

We select eight languages from Tatoeba, maintaining diversity in language families and scripts: Arabic, Mandarin Chinese, German, Finnish, French, Spanish, Vietnamese, and Turkish. The selection contains several high-resource languages (Arabic, Mandarin Chinese, German, French, and Spanish) for which strong performance is expected and in which failure to complete the task would be particularly striking, alongside languages with fewer resources (Finnish, Turkish, and Vietnamese).

\textbf{Distractors} \ We additionally generate synthetic data for our experiments. For each query in all of the datasets, we create a \textit{distractor} (a semantically similar text meant to distract evaluation) by prompting GPT-5.4-mini to paraphrase the query while preserving its meaning. These paraphrase queries serve as hard query-side negatives, testing whether models are capable of using task-specific instructions to shift the query embedding away from plain semantic similarity matching toward the actual task-specific target. Generation details and prompts are provided in Appendix~\ref{app:distractor-creation}.

\textbf{Models} \ \ The 8 models used in this study are 
Octen/Octen-Embedding-8B \citep{octen2025rteb}, 
codefuse-ai/F2LLM-v2-4B \citep{codefuseAI-f2llm-v2},
Qwen/Qwen3-Embedding-4B and 0.6B \citep{qwen3embedding},
microsoft/harrier-oss-v1-0.6b \citep{microsoft-harrier-oss},
BAAI/bge-m3 \citep{bge-m3},
intfloat/multilingual-e5-large-instruct \citep{wang2024multilingual-e5}, and
google/embeddinggemma-300m \citep{embedding_gemma_2025},
with Appendix \ref{app:embedding-models-table} providing further details. We aimed to select prominent embedding models with varying performance on the multilingual MMTEB benchmark \citep{enevoldsen-2025-mmteb}, and not limit our selection exclusively to the highest-scoring models. We also opted for variability in model size (in parameters) and embedding size (number of produced embedding dimensions), and lastly, omitted models with more than 8B parameters due to computational constraints. We use the \textit{sentence-transformers} library \citep{reimers-2019-sentence-bert-sentence-transformers}.

\textbf{Instructions} \ \ The full list of instructions we use in our experiments, the motivation for their selection, and other details are listed in Appendix \ref{app:instructions-prompts} with selected examples in Table \ref{tab:example-instructions}. We use 47 instructions (51 for Tatoeba: 4 additional instructions that mention the target language) in our first experiments and then narrow the selection to task-related instructions in the rest of the experiments. We intentionally include instructions for unrelated tasks, such as clustering, to test how the task-relevance of instruction influences the evaluation and to distinguish the varying ways different types of instructions can affect the resulting embeddings. 

\begin{table}[!ht]
\small
    \centering
    \caption{Examples of instructions in our experiments, with \{lang\} replaced with target language in Tatoeba evaluation. Instructions are intentionally varied. 
    }
    \label{tab:example-instructions}
    \begin{tabular}{cc}
    \toprule
    \textit{Retrieve parallel sentences.} &\textit{Retrieve the corresponding translation in \{lang\}.} \\
    \textit{Retrieve text based on user query.} & \textit{Retrieve semantically similar text.} \\
    \textit{Group documents related to the same topic.} &
    \textit{asdfjkl qpwoeiru zxcvbnm} \\  \bottomrule
    \end{tabular}
\end{table}

\subsection{Structural Statistics}
\label{sec:3-metrics}

We formulate the setting of retrieval and bitext mining and our metrics in the following way. Let $D_q$ be a collection of queries and $D_c$ the target documents (usually, \textit{corpus}), and $qrel: D_q \rightarrow D_c$ be a mapping between these sets marking relevance, more concretely, mapping a query $Q \in D_q$ to its correct target, $A \in D_c$. In our case, $qrel$ is always injective, but not necessarily surjective, and the relevance is always binary, meaning a target document either is or is not relevant to a given query. Distractors, $D$, are semantically related text for each $Q$. 
We denote adding the instruction $P$ to query $Q$ with $PQ$. 
Equations are given in Appendix \ref{app:formulas}.

\textbf{Similarity Improvement} \ \ We calculate similarity increase for an instruction $P$ over all $Q \in D_q$ for normalized embedding vectors as the difference of cosine similarities of $Q$ and $A$ compared to $PQ$ and $A$ (Eq. \ref{eq:sim}). If the value is greater than 0, adding the instruction $P$ increased the overall similarity to $A$; otherwise, it decreased the overall similarity. Similarity improvement is likely the most natural and intuitive way of viewing the effect of the instruction: a capable embedding model should react to a relevant instruction and move the query embedding closer to the correct target. However, naively moving towards the correct target may also mean moving towards incorrect targets.

\textbf{Displacement} \ \
We calculate the displacement caused by adding instruction $P$ to the query as the cosine distance between embeddings of $Q$ and $PQ$ (Eq. \ref{eq:dis}).
Values close to 0 indicate high similarity between $Q$ and $PQ$, meaning low displacement. We measure displacement to evaluate the magnitude of the effect adding instructions has: if $Q$ and $PQ$ are close, adding the instruction has only a limited effect. On the other hand, too much movement can be detrimental. We also calculate the displacement wrt. the targets $A$, and distractors $D$, separately, using the same formula, but substituting $emb(Q)$ with the embedding of the relevant target $emb(A)$, and $emb(PQ)$ with $emb(D)$, respectively.

\textbf{Angulation} \ \ Finally, we calculate angulation toward the correct answer to account for the drawback of displacement described above. We measure the cosine similarity between the vector from $emb(Q)$ to $emb(PQ)$, $\vec{v}_{Q,PQ}$, and the vector from $emb(Q)$ to $emb(A)$, $\vec{v}_{Q,A}$ (Eq. \ref{eq:ang}). Value 1 indicates that all movement was in the direction towards the target $A$, while 0 indicates orthogonal movement.

These three measures capture intuitive expectations on the effects the instructions have on the embeddings, and are intended to diagnose key characteristics related to our experiments.

\section{Results}
\label{sec:results}

\subsection{Instructions vs. Evaluation Performance}

\begin{table}[!hb]
\small
    \centering
    \caption{Evaluation of relevant against other instructions. Star ($^*$) indicates significance $p<0.05$.}
    \label{tab:relevant-evaluation}
    \begin{tabular}{@{}lcccccc@{}}
    \toprule
    & \multicolumn{3}{c}{ARCChallenge(ndcg@10)} & \multicolumn{3}{c}{Tatoeba(recall@1)} \\ \cmidrule(r){2-4} \cmidrule(l){5-7} 
    Model                                   & top-$k$       & $R^2$          & MV          & top-$k$      & $R^2$         & MV         \\ \midrule
    BAAI/bge-m3                             & 0.286         & 0.059       & 0.457*      & 0.350        & \textbf{0.065}     & 0.554*     \\
    codefuse-ai/F2LLM-v2-4B                 & \textbf{0.571}         & \textbf{0.272}       & \textbf{0.814}*      & 0.538        & 0.014     & 0.750*      \\
    google/embeddinggemma-300m              & 0.143         & 0.000       & -0.021      & 0.325        & 0.011     & 0.252*     \\
    intfloat/multilingual-e5-large-instruct & 0.429         & 0.082       & 0.571*      & 0.575        & 0.052     & 0.787*     \\
    microsoft/harrier-oss-v1-0.6b           & 0.429         & 0.103       & 0.543*      & 0.588        & 0.056     & 0.81*      \\
    Octen/Octen-Embedding-8B                & 0.429         & 0.053       & 0.536*      & \textbf{0.650}        & 0.024     & 0.779*     \\
    Qwen/Qwen3-Embedding-0.6B               & 0.429         & 0.092       & 0.600*      & \textbf{0.650}        & 0.054     & \textbf{0.821}*     \\
    Qwen/Qwen3-Embedding-4B                 & 0.143         & 0.069       & 0.436*      & 0.638        & 0.062     & 0.793*   \\\bottomrule
    \end{tabular}
\end{table}

We compare the evaluation scores of \textit{relevant} instructions (defined as instructions related to retrieval or bitext-mining, respectively, separately in each scenario, see Appendix \ref{app:instructions-prompts}) to those of other instructions with three methods: $R^2$ obtained for fitting a linear model on relevance-labels and scores, the effect-sizes of Mann-Whitney U-test (MW) (one-tailed, with the assumption that relevant instructions should yield higher scores), and by analyzing the fraction of relevant instructions in the top-$k$ scoring instructions (with $k$ defined as number of relevant instructions). For Tatoeba, which comprises 8 languages with differences in scores across one model, we use a linear mixed model, control for language, and report $R^2$ as the variance explained by the relevance label, making the value comparable to that of ARCChallenge. In the two other statistics, we normalize the evaluation scores before calculation to account for per-language variation.

The results are presented in Table \ref{tab:relevant-evaluation}. The fraction of relevant instructions in top-$k$ scoring instructions indicates that non-relevant instructions are found in top results in both datasets and all models, and often make up the majority. For example, even one of the models receiving the highest scores on ARC\-Challenge, Octen/Octen-Embedding-8B, has a keysmash instruction (see Table \ref{tab:example-instructions}) in the top results. Outside the clear outlier, codefuse-ai/F2LLM-v2-4B, the $R^2$-values are also consistently close to zero. On the other hand, the effect sizes of MW are large, which indicates that the distributions of relevant and other instructions are significantly different. Concretely, relevant instructions are always seen in the top of the results, but unrelated instructions regularly reach comparable performance. In conclusion, the relevance to task offers a distributional signal wrt.\ resulting performance in the standard embedding evaluation setup, but the dominant source of variation in scores is something other than intuitive and straightforward relatedness to given tasks, which aligns with related work. 

\subsection{Similarity Increase, Displacement and Angulation}

If the presumed quality and task-relatedness of the instruction does not fully explain the evaluation results, the most natural next step is to investigate how the instructions affect the embeddings. We analyze displacement, similarity increase, and angulation as defined in Section \ref{sec:3-metrics}. 

\begin{figure}[!t]
    \centering
    \begin{minipage}{\textwidth}
        \includegraphics[width=.31\linewidth]{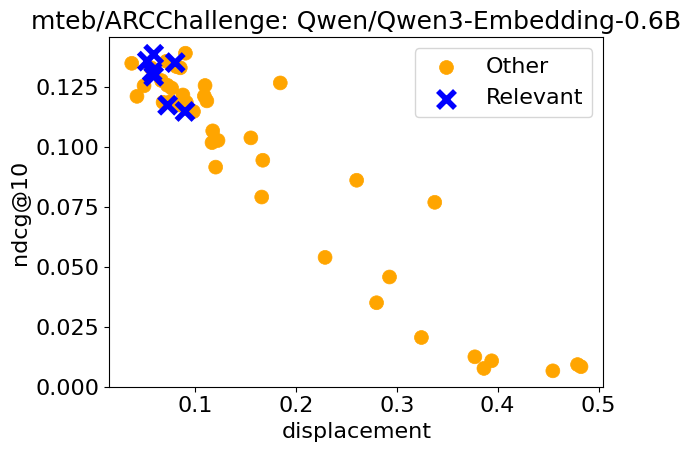}\hfill
        \includegraphics[width=.31\linewidth]{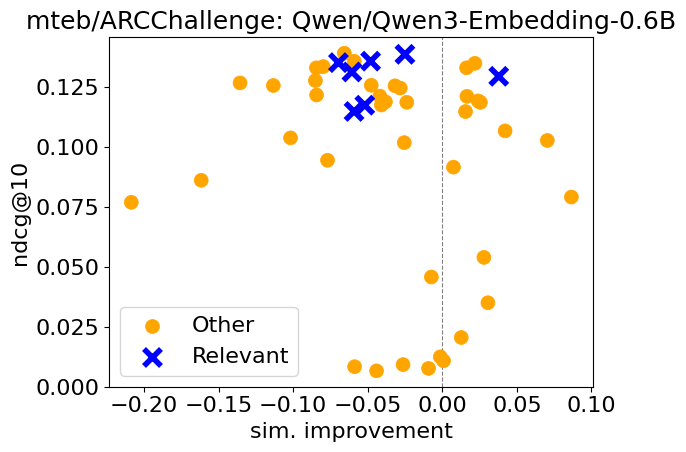}\hfill
        \includegraphics[width=.31\linewidth]{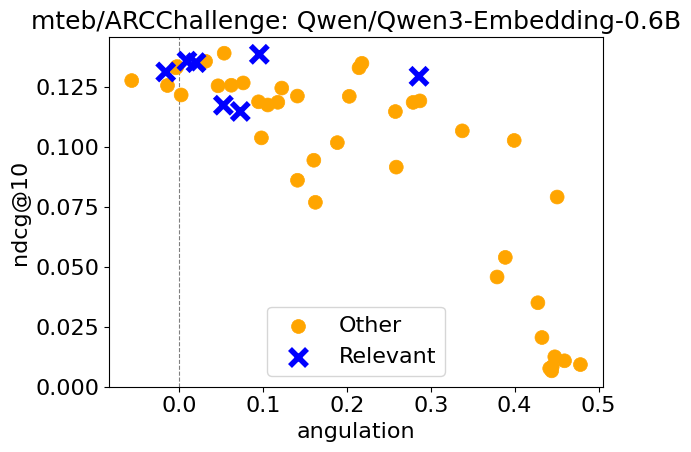}
    \end{minipage}
    \vspace{-0.1cm}\\
    \begin{minipage}{\textwidth}
        \includegraphics[width=.31\linewidth]{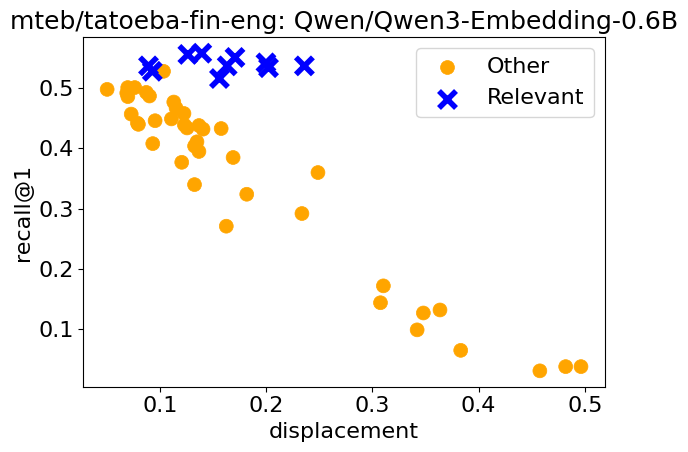}\hfill
        \includegraphics[width=.31\linewidth]{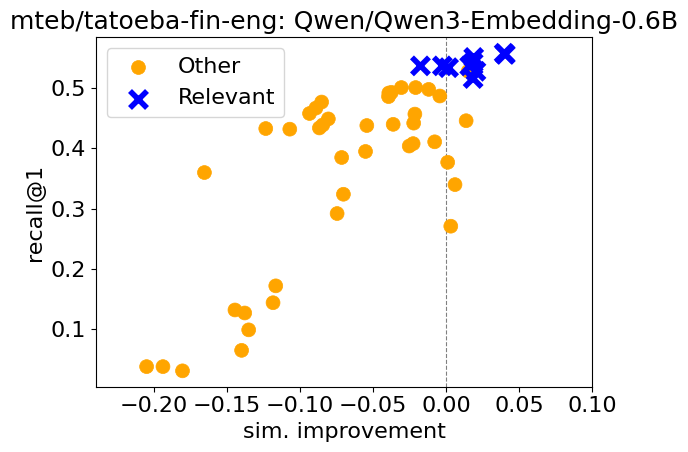}\hfill
        \includegraphics[width=.31\linewidth]{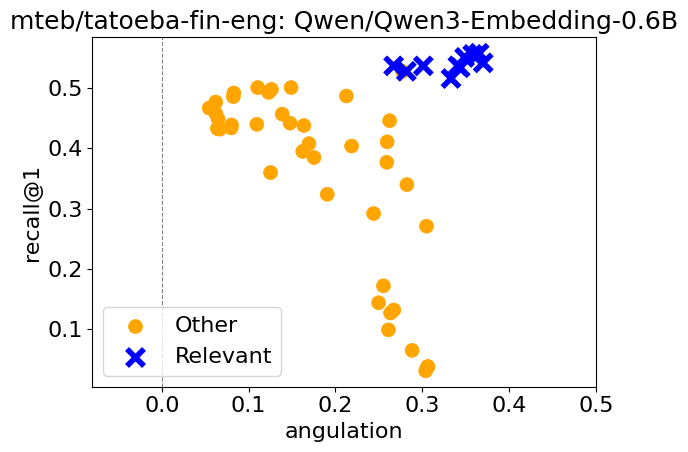}
        \caption{ARCChallenge (up) and Tatoeba:fin-eng (down), displacement, sim. improvement, and angulation. Each marker is a separate instruction. 
        On Tatoeba, relevant instructions are highly distinguishable, unlike ARCChallenge, where especially displacement is low for relevant instructions. 
        }
        \label{fig:3-metrics-on-ARC-and-tatoeba}
    \end{minipage}
\end{figure}

Figure \ref{fig:3-metrics-on-ARC-and-tatoeba} shows the results for Qwen/Qwen-Embedding-0.6b on ARCChallenge and Tatoeba. Clearly, overall low displacement correlates with performance, and the relevant instructions are not distinguishable from the others at the top end of the scores on ARCChallenge. Similarity improvement does not correlate with performance for relevant instructions, with the majority of retrieval instructions even found on the negative side. Angulation shows that the majority of the instructions receiving high scores also do not move along the query-target line. The same cannot be said for Tatoeba: relevant instructions display almost full separation from others. A relationship between less displacement and higher score is present, but its strength is pronouncedly different between relevant and other instructions. Relevant instructions display more angulation while retaining high performance. Similarity improvement for relevant instructions sits at approximately 0, and is negative for other instructions. Across other models, we see varying levels of dependence and separation of relevant and other instructions. To quantify these relationships for all models, we use cross-validated (N=200, with 5 stratified folds) ROC-AUC of logistic regression, with evaluation scores and each structural metric as predictors, to measure separation between relevant and other instructions, and additionally Spearman's $r$ to measure the monotonicity of the observed relationships. 

The results in Table \ref{tab:AUCs} match those of Figure \ref{fig:3-metrics-on-ARC-and-tatoeba}: values are lower for ARCChallenge compared to Tatoeba, which means the relevant instructions have more distinct behavior under the structural metrics on Tatoeba than ARCChallenge. The Spearman correlations, which demonstrate the monotonicity of the score vs.\ structural metric relationship, are presented in Table\ \ref{tab:spearmans-for-3-metrics}, and they display the dominance of similarity improvement on Tatoeba and the negative relationship between displacement and evaluation score for both datasets. 
Additionally, we visualize one clear outlier, codefuse-ai/F2LLM-v2-4B, the only model showing values above 0.9 on ARCChallenge, in Appendix\ \ref{app:outlier_results}.

\begin{table}[!h]
\centering
\small
\caption{Cross-validated ROC-AUC of logistic regression using the evaluation score and each structural metric (displacement (displ.), similarity improvement (sim.) and angulation (angl.)) as the predictors. High values indicate greater separability between relevant and other instructions, with 0.5 corresponding to chance level. Relevant instructions are more separable on Tatoeba.}
\label{tab:AUCs}
\begin{tabular}{@{}lcccccc@{}}
\toprule
                                        & \multicolumn{3}{c}{ARCChallenge (ndcg@10)}       & \multicolumn{3}{c}{Tatoeba (recall@1)}           \\ \cmidrule(r){2-4} \cmidrule(r){5-7}
Model                                   & displ.         & sim.           & angl.          & displ.         & sim.           & angl.          \\ \midrule
BAAI/bge-m3                             & 0.701          & 0.696          & 0.700          & \textbf{0.838} & \textbf{0.827} & \textbf{0.870} \\
codefuse-ai/F2LLM-v2-4B                 & \textbf{0.916} & \textbf{0.910} & \textbf{0.918} & \textbf{0.890} & \textbf{0.976} & \textbf{0.950} \\
google/embeddinggemma-300m              & 0.520          & 0.184          & 0.448          & 0.631          & 0.655          & 0.681          \\
intfloat/multilingual-e5-large-instruct & 0.784          & \textbf{0.800} & 0.769          & \textbf{0.877} & \textbf{0.999} & \textbf{1.000} \\
microsoft/harrier-oss-v1-0.6b           & 0.780          & \textbf{0.836} & \textbf{0.877} & \textbf{0.896} & \textbf{0.877} & \textbf{0.948} \\
Octen/Octen-Embedding-8B                & 0.742          & 0.707          & 0.723          & \textbf{0.942} & \textbf{0.883} & \textbf{0.999} \\
Qwen/Qwen3-Embedding-0.6B               & \textbf{0.843} & 0.746          & 0.776          & \textbf{0.958} & \textbf{0.880} & \textbf{0.998} \\
Qwen/Qwen3-Embedding-4B                 & 0.713          & 0.674          & 0.692          & \textbf{0.881} & \textbf{0.965} & \textbf{0.986} \\ \bottomrule
\end{tabular}
\end{table}

\begin{table}[!ht]
    \centering
    \small
    \caption{Spearman correlations between evaluation score and displacement (displ.), similarity improvement (sim.) and angulation (angl.). Values summarize the monotonicity seen in the results: the relationship between low displacement and high score is consistent in both datasets, while similarity improvement and angulation show more variation. Star ($^*$) indicates significance $p<0.05$.}
    \label{tab:spearmans-for-3-metrics}
\begin{tabular}{@{}lcccccc@{}}
\toprule
            & \multicolumn{3}{c}{ARCChallenge(ndcg@10)} & \multicolumn{3}{c}{Tatoeba(Recall@1)} \\ \cmidrule(r){2-4} \cmidrule(l){5-7} 
Model                                   & displ.       & sim.        & angl.        & displ.      & sim.       & angl.      \\ \midrule
BAAI/bge-m3                             & -\textbf{0.823}*      & \textbf{0.555}*      & -\textbf{0.763}*      & -\textbf{0.837}*     & \textbf{0.839}*     & -\textbf{0.510}*    \\
codefuse-ai/F2LLM-v2-4B                 & 0.207        & \textbf{0.884}*      & \textbf{0.522}*       & -\textbf{0.826}*     & \textbf{0.848}*     & 0.168*     \\
google/embeddinggemma-300m              & -\textbf{0.772}*      & 0.028       & -\textbf{0.542}*      & -\textbf{0.658}*     & \textbf{0.516}*     & -\textbf{0.503}*    \\
intfloat/multilingual-e5-large-instruct & -\textbf{0.639}*      & -0.132      & -0.369*      & -\textbf{0.699}*     & \textbf{0.864}*     & 0.118*     \\
microsoft/harrier-oss-v1-0.6b           & -\textbf{0.812}*      & 0.049       & -\textbf{0.514}*      & -0.323*     & \textbf{0.54}*      & 0.164*     \\
Octen/Octen-Embedding-8B                & -\textbf{0.587}*      & 0.496*      & -0.091       & -0.321*     & \textbf{0.695}*     & 0.251*     \\
Qwen/Qwen3-Embedding-0.6B               & -\textbf{0.845}*      & -0.265      & -\textbf{0.772}*      & -\textbf{0.609}*     & \textbf{0.792}*     & 0.215*     \\
Qwen/Qwen3-Embedding-4B                 & -\textbf{0.711}*      & -0.181      & -\textbf{0.672}*      & -\textbf{0.593}*     & \textbf{0.800}*     & 0.478*     \\ \bottomrule
\end{tabular}
\end{table}

\subsection{Why does this happen?}

The above results confirm that low displacement and small similarity gains show high results in standard evaluation. In Tatoeba, we saw a more desirable behavior: even if relevant instructions are not perfectly distinguishable by score, they have meaningfully different values in displacement, similarity improvement, and angulation compared to other instructions.
Our hypothesis is that current embedding fine-tuning and evaluation schemes do not incentivize models to react to instructions: in contrastive learning, in-batch and \emph{target-side} hard negatives may not sufficiently push instructed query embedding toward the answer region, especially when the evaluation corpus contains only plausible targets. In translation, mixing tasks during fine-tuning exposes the model to slightly more varied negatives, which may improve language sensitivity and explain the better results on Tatoeba.

\begin{figure}[!ht]
    \centering
    \includegraphics[width=0.48\linewidth]{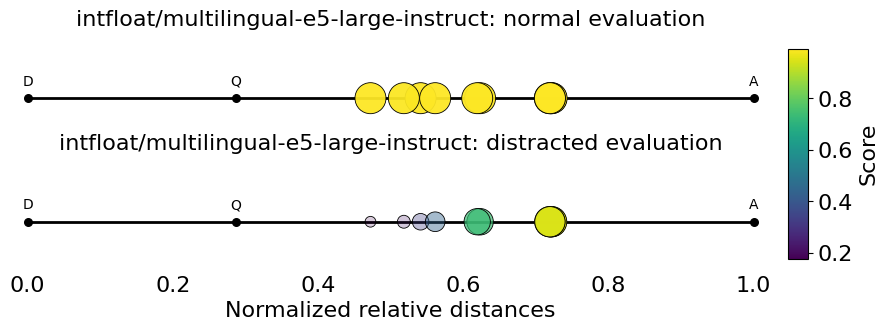}
     \includegraphics[width=0.48\linewidth]{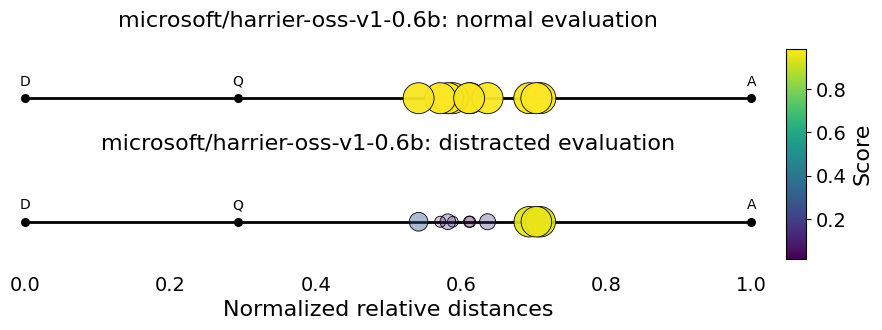}
    \caption{Taoteba:deu for multilingual-e5-large-instruct (left) and harrier-oss-v1-0.6b (right): Evaluation without (up) and with distractors (down). Each circle corresponds to a relevant instruction-query, and its position indicates its relative normalized displacement between Q=query, D=distractor, and A=target. The more movement towards the answer, the better the score is retained.}
    \label{fig:lineplot-with-tatoeba-deu}
\end{figure}

To test our assumption, we introduce distractors to the retrieval pool to probe whether the instructions can survive the added text that is related semantically but not according to the instruction. For example, in ARCChallenge, the instruction to retrieve the answer should result in retrieving the answer, and not a related question. We evaluate with Recall$@$1 to pinpoint the effect of one distractor. Results of this evaluation are in Table\ \ref{tab:distractor-eval-results}, where we show the maximal performance over all instructions (and all languages in Tatoeba). They show that the addition of distractors completely destroys the performance for all instructions on ARCChallenge. Related retrieval instructions are also generally worse. On Tatoeba, two models almost perfectly retain the performance, and the best-scoring instructions are relevant. Figure \ref{fig:lineplot-with-tatoeba-deu} illustrates the effect of distractors with these two models and relevant instructions with respect to placement between distractor, query, instructed query, and target, summarizing aspects of displacement, similarity improvement and angulation into one figure: clearly, relatively more movement away from query and more towards the target help retain performance when distractors are added. 
From these results, we see that models that sufficiently react to instructions exist, but the majority fail to do so. In cases where distractors do not destroy the performance, the fraction of movement should be more on the target side. Next, we experiment with fine-tuning the model to explicitly have this characteristic and successfully respond to instructions.

\begin{table}[!t]
\small
\centering
\caption{Distractor results for ARCChallenge and Tatoeba, both with maximal recall@1 over all instructions (and all languages in Tatoeba): normal column refers to normal evaluation setup with no distractors, distr. column to setup where distractors are introduced to the target corpus, and rel. showing the best performance of relevant instructions, if lower than the distr. column. }
\label{tab:distractor-eval-results}
\begin{tabular}{@{}llllllll@{}}
\toprule
    & \multicolumn{3}{c}{ARCChallenge }                 &  & \multicolumn{3}{c}{Tatoeba (max)}             \\ \cmidrule(lr){2-4} \cmidrule(l){6-8} 
Model                                   & normal & distr. & rel.&  & normal        & distr. & rel.\\ \midrule
BAAI/bge-m3                             & 0.032   & 0.001          & 0.000                   &  & 0.987          & 0.471          & --                 \\
codefuse-ai/F2LLM-v2-4B                 & 0.214   & 0.062          & --                 &  & 0.998          & 0.626          & 0.55                  \\
google/embeddinggemma-300m              & 0.019   & 0.005          & 0.002                 &  & 0.725          & 0.25           & --                  \\
intfloat/multilingual-e5-large-instruct & 0.074   & 0.007          & 0.0                   &  & \textbf{0.994} & \textbf{0.954} & --        \\
microsoft/harrier-oss-v1-0.6b           & 0.106   & 0.007          & --                &  & \textbf{0.988} & \textbf{0.956} & --       \\
Octen/Octen-Embedding-8B                & 0.149   & 0.005          & 0.001                 &  & 0.992          & 0.671          & --                 \\
Qwen/Qwen3-Embedding-0.6B               & 0.059   & 0.007          & 0.0                   &  & 0.977          & 0.500            &--                   \\
Qwen/Qwen3-Embedding-4B                 & 0.112   & 0.038          & --                 &  & 0.994          & 0.529          & --                \\ \bottomrule
\end{tabular}
\end{table}

\section{Distractor Fine-tuning}

To test whether distractor robustness can be improved by including such examples during model training, we fine-tune the Qwen/Qwen3-Embedding-0.6B model, for which all retrieval prompts were affected by added distractors. We fine-tune separate models for QA and bitext mining, with the latter fine-tuned multilingually on a mixture of all eight languages. Fine-tuning is performed using the SWIFT framework \citep{zhao2024swiftascalablelightweightinfrastructure}. Details of the dataset splits and sizes, data preprocessing, and fine-tuning parameters are available in Appendix~\ref{app:fine-tuning-details}.

We use SQuAD.v1 for QA fine-tuning, since ARCChallenge contains only a test set, and obtain Tatoeba training examples from the OPUS dataset \citep{tiedemann-2012-parallel}. We use the same training setup for both tasks. Each example consists of a query ($Q$), its retrieval target ($A$), and a distractor ($D$), which is a generated paraphrase of $Q$. For QA, $Q$ is a question and $A$ its answer paragraph, while for bitext mining, $Q$ is an English sentence and $A$ its target-language translation.
To avoid catastrophic forgetting, we fine-tune jointly on retrieval (QA, bitext mining) and semantic textual similarity (STS), constructing $P_{task}Q \rightarrow A$ for the retrieval tasks and $P_{sts}Q \rightarrow D$ for STS, where $P_{task}$ is $P_{qa}$ or $P_{bitext}$. Thus, $D$ serves as a positive target for STS but a query-side negative for the retrieval tasks. Each positive target is accompanied by eight random target-side negatives. For 50\% of the training examples, one negative is replaced by the wrong-task target: query-side distractor $D$ for the retrieval tasks and $A$ for STS. QA and bitext models are fine-tuned separately, but for bitext mining, examples from all eight languages are pooled with the language specified in instructions.

\subsection{Evaluation and Results}

We evaluate the fine-tuned models under three settings: \emph{Normal}, \emph{Random distractors}, and \emph{Paraphrase distractors}. \emph{Normal} measures standard retrieval performance without added distractors. \emph{Random} adds random query-side examples as distractors (other questions for QA, and other English sentences for bitext mining), allowing us to analyze both the effect of increased candidate set and whether arbitrary query-side distractors degrade retrieval performance. \emph{Paraphrase} adds query-side distractors that are semantically highly similar, but do not fulfill the task instructions, testing the model's ability to follow instructions instead of plain semantic similarity. In both settings, one distractor is added per query, increasing the candidate pool size by the total number of queries.

\begin{figure}
    \centering
     \begin{minipage}[t]{0.45\linewidth}
     \vspace{0pt}
    \includegraphics[width=\linewidth]{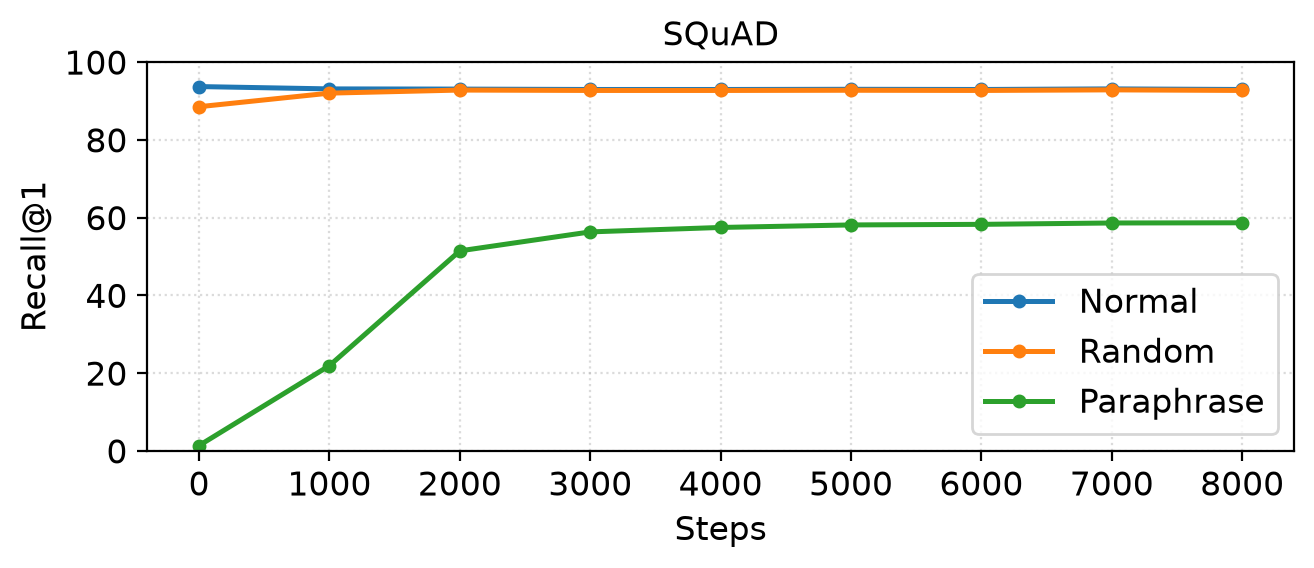}
    \captionof{figure}{SQuAD development performance across fine-tuning steps.}
    \label{fig:finetuning-curve-squad}
    \end{minipage}
    \hfill
    \begin{minipage}[t]{0.45\linewidth}
    \vspace{0pt}
   \includegraphics[width=\linewidth]{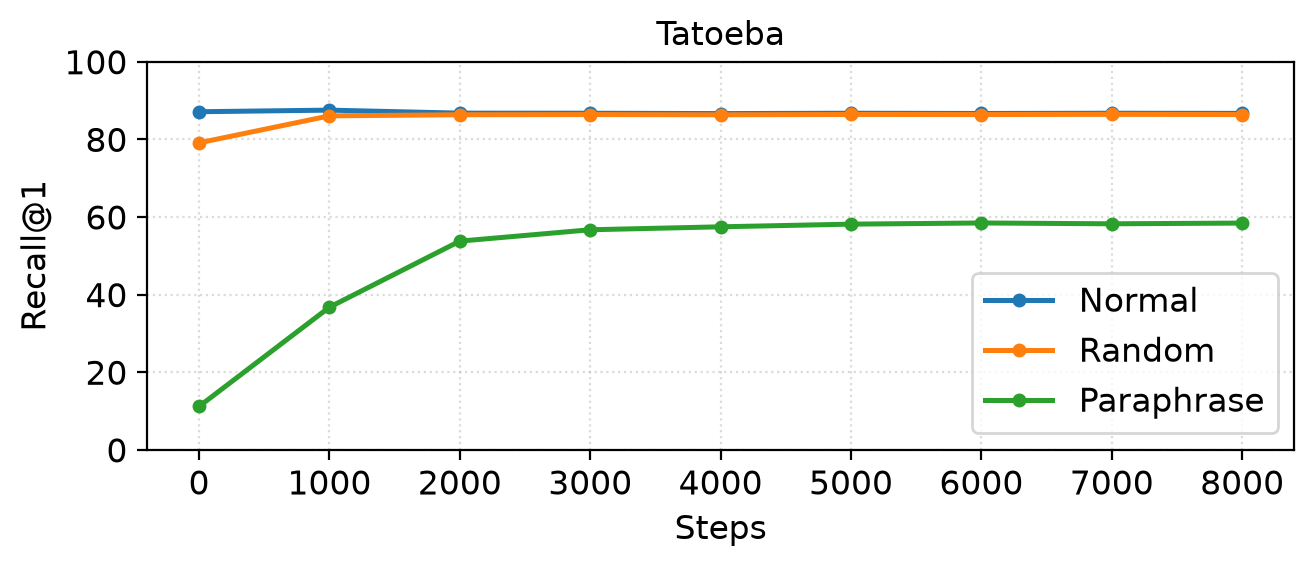}
    \captionof{figure}{Tatoeba development performance across fine-tuning steps.}
    \label{fig:finetuning-curve-tatoeba}
    \end{minipage}
\end{figure}

\begin{table}[]
\centering
\small
\caption{Recall@1 and Recall@5 test results on SQuAD and Tatoeba for the baseline Qwen/Qwen3-Embedding-0.6B compared to fine-tuned models. Tatoeba results are reported as a mean across 8 languages (per language results in Appendix \ref{app:tatoeba-results}).}
\label{tab:main-finetuning-results}
\begin{tabular}{@{}lllllllll@{}}
\toprule
   & & \multicolumn{3}{c}{SQuAD}  &  & \multicolumn{3}{c}{Tatoeba (mean)}  \\ \cmidrule(lr){3-5} \cmidrule(l){7-9} 
Metric &Model    & normal & random & paraphr.&  & normal   & random & paraphr.\\ \midrule

Recall@1 & baseline           & 0.724 & 0.685  & 0.014  &   & 0.844 & 0.769 & 0.102 \\
  &fine-tuned         & 0.706 & 0.702 & 0.470   &   & 0.837 & 0.833 & 0.569  \\
Recall@5 &baseline           & 0.918 & 0.892 & 0.778   &   & 0.926 & 0.878 & 0.832  \\
  & fine-tuned         & 0.912 & 0.910 & 0.890   &   & 0.923 & 0.921 & 0.908  \\\bottomrule
\end{tabular}
\end{table}

Figures~\ref{fig:finetuning-curve-squad} and \ref{fig:finetuning-curve-tatoeba} show SQuAD and Tatoeba development performance over the fine-tuning steps. Fine-tuning substantially improves performance in the distractor setting, while performance in the normal setting changes only slightly. Most of the improvement occurs within the first 5,000 steps (approx. 30\% of one epoch, 140K/150K training examples in total), after which development performance plateaus. 
The final test-set results are reported in Table~\ref{tab:main-finetuning-results} for the baseline Qwen/Qwen3-Embedding-0.6B model and its fine-tuned variants. A similar trend is observed for both SQuAD and Tatoeba. For the baseline model, \emph{paraphrase distractors} almost completely destroy Recall@1 performance, despite being instructed to retrieve the target, the model typically retrieves the query-side distractor instead. By contrast, \emph{random distractors} cause only a moderate drop in performance, indicating that simply increasing the candidate set or adding unrelated queries does not substantially mislead the model. A similar pattern is visible for Recall@5, where adding a single semantically similar paraphrase distractor per query does not always push the correct target out of the top five. This highlights the need for semantically similar query-side distractors in evaluation. 

The fine-tuned models show different behavior when distractors are added. Their performance is nearly unaffected by random distractors, and although paraphrase distractors still reduce performance, the degradation is substantially smaller than for the baseline. This suggests that the model can learn 
to follow task-specific instructions from a relatively small amount of fine-tuning data. Importantly, this improvement does not substantially degrade standard retrieval. The \emph{Normal} results in Table~\ref{tab:main-finetuning-results} remain close to the baseline. In Appendix~\ref{app:finetuning-wo-hardnegatives}, we show that the improved robustness to paraphrase distractors is not due to task-specific fine-tuning alone, but specifically to the addition of query-side negatives. When the models are fine-tuned on the same data (SQuAD, Tatoeba) using the same setup but excluding query-side negatives, performance under paraphrase distractors remains flat, and the gap between the \emph{Normal} and \emph{Random} evaluation settings remains.

\begin{figure}
    \centering
    \includegraphics[width=0.75\linewidth]{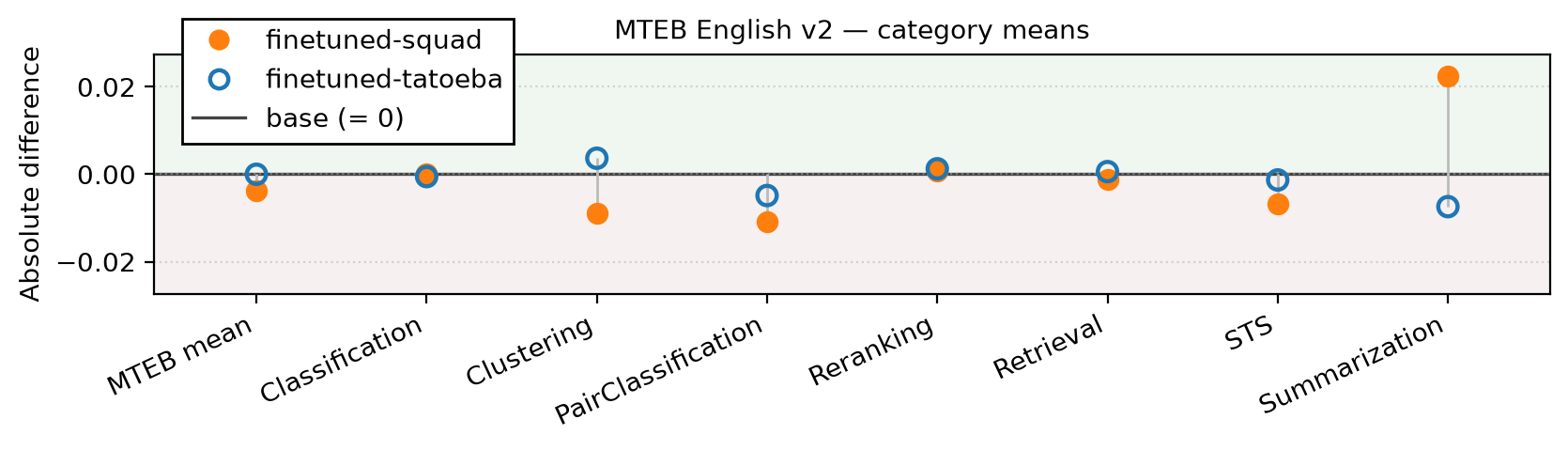}
    \caption{Performance on the MMTEB benchmark categories for fine-tuned models. The absolute differences are small, meaning our fine-tuning setup does not affect the overall performance. }
    \label{fig:mteb-category-mean}
\end{figure}

\subsection{After Fine-tuning: What happened?}

Fine-tuning greatly increased the performance with distractors. Next, we shortly analyze the effects the fine-tuning had on the structural metrics introduced in Section~\ref{sec:3-metrics} and out-of-domain tasks.
In the SQuAD-finetuned model, similarity improvement increases: on average, after finetuning, all relevant instructions increase similarity to the answer more than before finetuning, not only the finetuning instruction. The effect is opposite for Tatoeba: the similarity improvement decreases slightly in all languages, but less for the relevant instructions.
Displacement grows in both finetuned models.
Both models also increase in angulation, and especially on Tatoeba, most increase is seen in the prompt where the language is explicitly mentioned, which is a desirable outcome of the fine-tuning.

\begin{figure}
    \centering
    \begin{minipage}{.48\textwidth}
    \centering
        \includegraphics[width=\linewidth]{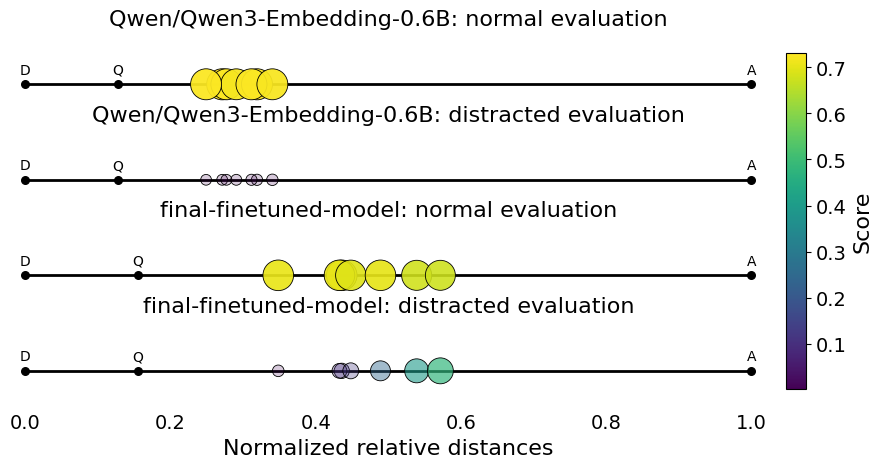}
        \caption{SQuAD: before and after for relevant instructions, with and without distractors.}
        \label{fig:squad_before_and_after}
    \end{minipage}
    \ 
    \begin{minipage}{.48\textwidth}
    \centering
        \includegraphics[width=\linewidth]{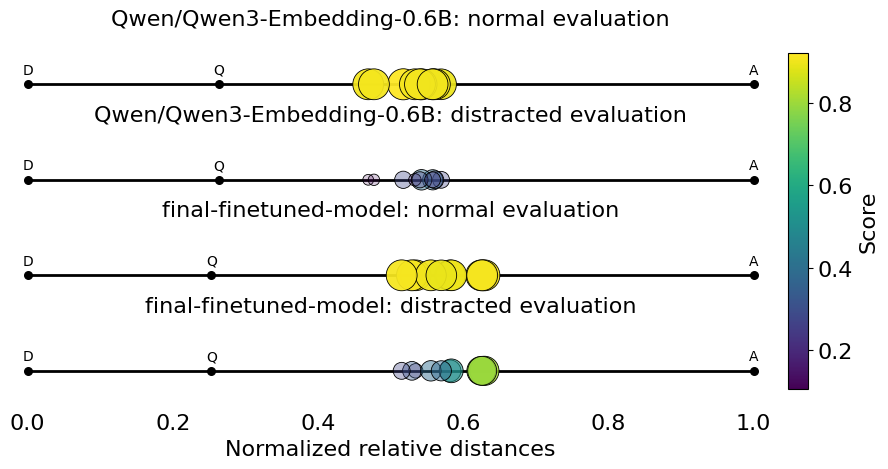}
        \caption{Tatoeba:fra: before and after for relevant instructions, with and without distractors.}
        \label{fig:tatoeba_before_and_after}
    \end{minipage}
    \label{fig:placeholder}
\end{figure}

Figures \ref{fig:squad_before_and_after} and \ref{fig:tatoeba_before_and_after} illustrate the changes with relative distance plots. Fine-tuning successfully yielded characteristics identified as beneficial in Section \ref{sec:results}: the instructions' effect is larger, and the performance is retained better when distractors are added to the retrieval pool. 

\textbf{Out-of-domain performance} \ \ We also ran the full MTEB English v2 evaluation \citep{enevoldsen-2025-mmteb} on the finetuned models, the results are shown in Figure \ref{fig:mteb-category-mean}, averaged per task category, and in Appendix \ref{app:full-mteb-results} in full. They show that while the fine-tuning for distractors has effects on the results on other benchmarks, the differences are small and distractor fine-tuning does not destroy the performance on other tasks. We also evaluated the SQuAD-fine-tuned model on ARCChallenge, see Appendix \ref{app:outlier_results} Table \ref{tab:after_finetuning_arcchallenge}. While the baseline performance is already low, fine-tuning manages to make the paraphrase distracted evaluation slightly improve for Recall@1 and Recall@5.

\section{Discussion and Conclusion}

In this paper, we hypothesized that current training and evaluation setups may overestimate the robustness of instructed text embedding models, and may not fully capture challenges encountered in real-world retrieval settings, particularly if candidates contain semantically similar query-side distractors. We investigated this claim from multiple angles: how instruction relevance is tied to performance, how simple structural measures show the effect of instructions, how introducing distractors negatively affects the models, and how fine-tuning with carefully selected negatives can mitigate this effect substantially. Our findings align with prior work showing that instructions have complex and poorly understood effects on embedding models, while extending this line of research toward geometric interpretations of representational change. Beyond analysis, we introduce both a probing framework for assessing inadequate instruction-following and a fine-tuning strategy to address it. Taken together, our results suggest that the standard evaluation paradigm may overestimate genuine instruction sensitivity. We therefore argue for a shift toward evaluation settings that pool across tasks, exposing models to instruction-sensitive, multi-task retrieval challenges. 

The main limitation of our study is the number of tasks investigated. We focus on retrieval and bitext-mining, and leave out other common embedding tasks, such as clustering. Similarly, while we study 8 prominent text embedding models, we cannot guarantee that the results generalize beyond this set. However, the value of the introduced methodology does not depend on universality, and the ability to identify and correct inadequate instruction-following behavior and the fine-tuning framework remains applicable to any model exhibiting similar characteristics. To reduce sensitivity to instruction selection, we deliberately use a large and varying instruction set with varying quality. The synthetic distractor data is generated using a single model and we do not ablate over different model choices. However, because paraphrasing of short queries imposes relatively constrained generation, we expect the choice of generation model to have only a limited effect on the results. In the future, we will extend the analysis to cover more task categories and query--target relations.

\subsection*{Acknowledgments and Disclosure of Funding}

This research has been supported by Finland's Ministry of Education and Culture’s Doctoral Education Pilot under Decision No. VN/3137/2024-OKM-6 (The Finnish Doctoral Program Network in Artificial Intelligence, AI-DOC); the Digital Europe Programme under grant agreement No. 101195233 (OpenEuroLLM); the Human Diversity consortium, under the Profi7 program (grant no. 352727) and the Centre of Excellence (grant no. 374221) by the Research Council of Finland; the Research Council of Finland through FIN-CLARIAH research infrastructure (project 358720, which has also received funding from the European Union -- NextGenerationEU instrument), and the Mechanisms of Register Variation in Massively Multilingual Web-Scale Corpora (project 362459). This project is funded by the European Union. Views and opinions expressed are however those of the author(s) only and do not necessarily reflect those of the European Union or the European Education and Culture Executive Agency (EACEA). Neither the European Union nor EACEA can be held responsible for them.
We thank CSC -- IT Center for Science, Finland, for computational resources.

\bibliography{refs}
\bibliographystyle{iclr2027_conference}

\appendix
\input{appendix}

\end{document}

%% file: math_commands.tex
\usepackage{amsmath,amsfonts,bm}

\def\eqref#1{equation~\ref{#1}}

\def\1{\bm{1}}

\DeclareMathAlphabet{\mathsfit}{\encodingdefault}{\sfdefault}{m}{sl}
\SetMathAlphabet{\mathsfit}{bold}{\encodingdefault}{\sfdefault}{bx}{n}



%% file: appendix.tex
\section{Generating Query-Side Distractors}
\label{app:distractor-creation}

For each query used in our experiments, we generated exactly one paraphrase to serve as query-side distractors. This totals to 185,254 generated paraphrases. We used OpenAI's GPT-5.4-mini (version 2026-03-17) through their Batch API. The total cost of the paraphrase generation is \$10.22, the majority accounting for the fine-tuning data.

Generation prompts are provided in Table \ref{tab:data-generation-prompts}, with \{input\} replaced with the actual text.

\begin{table}[b!]
    \centering
    \caption{Prompts used to generate synthetic distractors.}
    \begin{tabular}{lp{0.72\linewidth}}
    \hline
    \textbf{Dataset} & \textbf{Prompt} \\\hline
      \textbf{SQuAD} &  Paraphrase the following question while preserving its meaning exactly. Use different wording and avoid producing a near-identical phrasing.
\par\vspace{0.5em}
Output only the paraphrased question and nothing else.
\par\vspace{0.5em}
Question: \{input\}   \\\\
      \textbf{Tatoeba} & Paraphrase the following text while preserving its meaning exactly. Use different wording and avoid producing a near-identical phrasing.
\par\vspace{0.5em}
Output only the paraphrased text and nothing else.
\par\vspace{0.5em}
Text: \{input\} \\\\
      \textbf{ARCChallenge} & Paraphrase the following text while preserving its meaning exactly. Use different wording and avoid producing a near-identical phrasing.
      \par\vspace{0.5em}
      Output only the paraphrased text and nothing else.
      \par\vspace{0.5em}
      Text: \{input\} \\\hline
  
    \end{tabular}
    \label{tab:data-generation-prompts}
\end{table}

\section{Embedding Models}
\label{app:embedding-models-table}
Table \ref{tab:models} lists the number of parameters and dimensionality of the output embedding for each model. The licences for the models are

\textbf{MIT}: BAAI/bge-m3 , intfloat/multiligual-e5-large-instruct, microsoft/harrier-oss-v1-0.6b

\textbf{Apache-2.0}: Qwen/Qwen3-Embedding-4B, Qwen/Qwen3-Embedding-0.6B,  Octen/Octen-Embed\-ding-8B, codefuse-ai/F2LLM-v2-4B

\textbf{Gemma Licence}: google/embeddinggemma-300m.

\begin{table}[!ht]
\small
\centering
\caption{Models used in this study.}
\label{tab:models}
\begin{tabular}{@{}llll@{}}
\toprule
Model    & Parameters & Dimensions &  Citation \\ \midrule
Octen/Octen-Embedding-8B    &  7.6B &  4096 &\citet{octen2025rteb} \\
codefuse-ai/F2LLM-v2-4B    & 4B   &  2560& \citet{codefuseAI-f2llm-v2} \\
Qwen/Qwen3-Embedding-4B    &  4B & 2560&   \citet{qwen3embedding}  \\
Qwen/Qwen3-Embedding-0.6B & 596M & 1024& \citet{qwen3embedding} \\
microsoft/harrier-oss-v1-0.6b & 596M & 1024 & \citet{microsoft-harrier-oss}  \\
BAAI/bge-m3 & 568M & 1024& \citet{bge-m3} \\
intfloat/multilingual-e5-large-instruct & 560M & 1024 & \citet{wang2024multilingual-e5} \\ 
google/embeddinggemma-300m & 308M & 768 &\citet{embedding_gemma_2025} \\ \bottomrule
\end{tabular}
\end{table}

\section{Instructions}
\label{app:instructions-prompts}

We list all 47 (51 for bitext-mining) instructions in Table \ref{tab:all_prompts}. We use the instruction template \textit{Instruct: $<$Instruction$>$ \textbackslash nQuery: $<$Query$>$}. Additionally, we also calculate the query embeddings without instructions (or the template), making the total number of dataset embedding iterations 48/52. The instruction selection was started by searching for task-specific instructions, which are given for certain datasets, such as ARCChallenge. Next, we included the MTEB ``abstask'' instructions, instructions defined per task category in the MTEB library. We then augmented this selection with 1. task-specific instructions for unrelated tasks, 2. Vague instructions, written by authors, and 3. two to five LLM-generated instructions per category (Claude-Sonnet-4.6, with prompt ``Generate 10 instructions of varying quality for MTEB-style \{task category name\} embedding evaluation''), out of which we selected 2-5 with the least overlap with the instructions selected in the previous steps to diversify the selection. For bitext-mining, 4 instructions are included with and without the language specified in the instruction; for retrieval, only the instructions without language specification are included (concretely: ``in \{lang\}'' omitted). This makes bitext-mining have 4 additional instructions.

\begin{table}[]
\scriptsize
\centering
\begin{tabular}{@{}p{5cm}p{3.5cm}ll@{}}
\toprule
Instruction                                                                                              & Source                                                                                                                    & Motivation            & Relevant              \\ \midrule
"Identify categories in user   passages."                                                                & MTEB, clustering                                                                                                          & MTEB abstask         &                   \\
"Classify   user passages."                                                                              & MTEB, classification                                                                                                      & MTEB abstask         &                   \\
"Retrieve   text that are semantically similar to the given text."                                       & MTEB, pair classification                                                                                                       & MTEB abstask         &                   \\
"Retrieve   text based on user query."                                                                   & MTEB, retrieval                                                                                                           & MTEB abstask         & Retrieval         \\
"Retrieve   semantically similar text."                                                                  & MTEB, STS                                                                                                                 & MTEB abstask         &                   \\
"Given a   news summary, retrieve other semantically similar summaries."                                 & MTEB, summarization                                                                                                             & MTEB abstask         &                   \\
"Retrieve   parallel sentences."                                                                         & MTEB, bitext-mining                                                                                                           & MTEB abstask         & Bitext-mining     \\
"Retrieve   the answer to the question.",                                                                & MTEB, ARCChallenge                                                                                                                 & Task-specific         & Retrieval         \\
"Given a   question, retrieve the passage that best answers it.",                                         & Generated+Authors                                                                                                         & Retrieval, diversity  & Retrieval         \\
"Retrieve",                                                                                              & Generated+Authors                                                                                                         & Retrieval, diversity  & Retrieval         \\
"Find the   most relevant passage that directly answers the question.",                                   & Generated+Authors                                                                                                         & Retrieval, diversity  & Retrieval         \\
"Given a   question, find a related document."                                                            & Generated+Authors                                                                                                         & Retrieval, diversity  & Retrieval         \\
f"Retrieve   the corresponding translation in \{lang\}.",                                                  & Generated+Authors                                                                                                         & Bitext, diveristy     & Bitext-mining x2  \\
f"Given an   English sentence, find its translation in \{lang\}.",                                         & Generated+Authors                                                                                                         & Bitext, diveristy     & Bitext-mining x2  \\
f"Retrieve   parallel sentences in \{lang\}",                                                              & Generated+Authors                                                                                                         & Bitext, diveristy     & Bitext-mining x2  \\
f"Translate   to \{lang\}.",                                                                               & Generated+Authors                                                                                                         & Bitext, diveristy     & Bitext-mining x2  \\
"Find a   sentence that has similar meaning.",                                                           & Generated+Authors                                                                                                         & Bitext, diveristy     & Bitext-mining     \\
"Summarize   the given paragraph into a short paragraph.",                                               & Generated+Authors                                                                                                         & Summ., diversity      &                   \\
"Match the   text to the short abstract.",                                                               & Generated+Authors                                                                                                         & Summ., diversity      &                   \\
"Retrieve a   related summary."                                                                          & Generated+Authors                                                                                                         & Summ., diversity      &                   \\
"Classify   texts as entailment or contradiction."                                                       & Generated+Authors                                                                                                         & Pair clf, diverisity  &                   \\
"Determine   whether the following two passages are contradictory, entailed, or   neutral."              & Generated+Authors                                                                                                         & Pair clf, diverisity  &                   \\
"Classify   based on semantic similarity."                                                               & Generated+Authors                                                                                                         & Pair clf, diverisity  &                   \\
"Classify   the following text into a meaningful cluster based on its content.",                         & Generated+Authors                                                                                                         & Classif., diversity   &                   \\
"Group   documents related to the same topic.",                                                          & Generated+Authors                                                                                                         & \multicolumn{2}{l}{Clustering, diversity} \\
"Cluster."                                                                                               & Generated+Authors                                                                                                         & \multicolumn{2}{l}{Clustering, diversity} \\
"Retrieve a   similar passage.",                                                                         & Generated+Authors                                                                                                         & STS, diversity        &                   \\
"Represent   this sentence for a natural language understanding task.",                                  & Generated+Authors                                                                                                         & STS, diversity        &                   \\
"Group   passages based on semantic similarity."                                                         & Generated+Authors                                                                                                         & STS, diversity        &                   \\
"Retrieve   duplicate questions from the forum.",                                                        & MTEB, \href{https://huggingface.co/datasets/mteb/sprintduplicatequestions-pairclassification}{Sprint-duplicate-questions} & Task-specific         &                   \\
"Retrieve   the most relevant problem description for the given code   implementation."                  & MTEB, \href{https://huggingface.co/datasets/mteb/coreb-c2c-retrieval}{coreb-retrieval}                                    & Task-specific         &                   \\
"Rerank   products by relevance to the e-commerce query.",                                               & MTEB, \href{https://huggingface.co/datasets/thebajajra/eress}{E-commerce reranking}                                                                                                & Task-specific         &                   \\
"Classify   the sentiment of the given review as positive or negative.",                                  & MTEB, \href{https://huggingface.co/datasets/mteb/amazon_reviews_multi}{Amazon-review-multi}                               & Task-specific         &                   \\
"Identify   the topic or theme of Reddit posts based on the titles."                                      & MTEB, \href{https://huggingface.co/datasets/mteb/reddit-clustering}{reddit-clustering}                                    & Task-specific         &                   \\
"Identify   the main and secondary category of Arxiv papers based on the titles."                         & MTEB, \href{https://huggingface.co/datasets/mteb/arxiv-clustering-s2s}{arxiv-clustering}                                  & Task-specific         &                   \\
"Generate a   creative story based on the prompt.",                                                       & Generated+Authors                                                                                                         & \multicolumn{2}{l}{No corresponding task} \\
"Extract   all named entities from the text and categorise them as PERSON, LOCATION, or   ORGANIZATION.", & Generated+Authors                                                                                                         & \multicolumn{2}{l}{No corresponding task} \\
"Search for related information.",                                                                      & Generated+Authors                                                                                                         & Vague                 &                   \\
"Find documents related to the topic.",                                                                 & Generated+Authors                                                                                                         & Vague                 &                   \\
"Find something useful.",                                                                               & Generated+Authors                                                                                                         & Vague                 &                   \\
"Retrieve the passage that contains overlapping vocabulary with the source   sentence.",                & Generated+Authors                                                                                                         & Vague                 &                   \\
"asdfjkl qpwoeiru zxcvbnm",                                                                            & Authors                                                                                                                   & Vague                 &                   \\
"hgJKSbf oiawnef LKJHDS kdjfbs",                                                                       & Authors                                                                                                                   & Vague                 &                   \\
"!!!! ???   \#\#\# @@@",                                                                                    & Authors                                                                                                                   & Vague                 &                   \\
"Given a   question, retrieve Wikipedia passages that answer the question."                              & Authors                                                                                                                   & Finetuning            & Retrieval         \\
"Given a   question, retrieve questions that are semantically equivalent to the given   question."       & Authors                                                                                                                   & Finetuning            &                   \\ \bottomrule
\end{tabular}
\caption{Instructions used in our experiments, their sources, reasoning for their selection, and which task they are considered to be relevant.}
\label{tab:all_prompts}
\end{table}

\section{Formulas for structural metrics}
\label{app:formulas}
The formulas we use to calculate similarity improvement, displacement, and angulation are presented in Equations \ref{eq:sim}--\ref{eq:ang}. The embedding of a text is represented as $emb(\cdot)$, cosine similarity as $\cos(\cdot, \cdot)$, $A$ represents the relevant target for each query $Q$ in the collection of queries $D_q$, and $PQ$ represents the combination of instruction $P$ and query $Q$. The vector from $emb(Q)$ to $emb(PQ)$ is $\vec{v}_{Q,PQ}$, and the vector from $emb(Q)$ to $emb(A)$ is $\vec{v}_{Q,A}$.

\begin{equation}
\label{eq:sim}
    \text{Similarity Impr.}(P) = \frac{1}{|D_q|}\sum_{Q \in D_q} \cos(emb(PQ), emb(A)) - \cos(emb(Q), emb(A)),
\end{equation}
\begin{equation}
\label{eq:dis}
\text{Displacement}(P) = \frac{1}{|D_q|} \sum_{Q \in D_q} 1-\cos(emb(PQ), emb(Q)).
\end{equation}
\begin{equation}
\label{eq:ang}
    \text{Angulation}(P) = \frac{1}{|D_q|} \sum_{Q \in D_q} \cos(\vec{v}_{Q,PQ}, \vec{v}_{Q,A}).
\end{equation}

\section{Supporting results}
\label{app:outlier_results}

\begin{figure}
    \centering
    \includegraphics[width=0.31\linewidth]{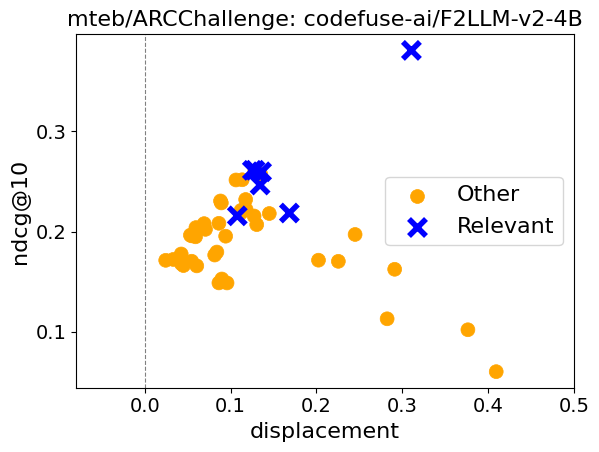}
     \includegraphics[width=0.33\linewidth]{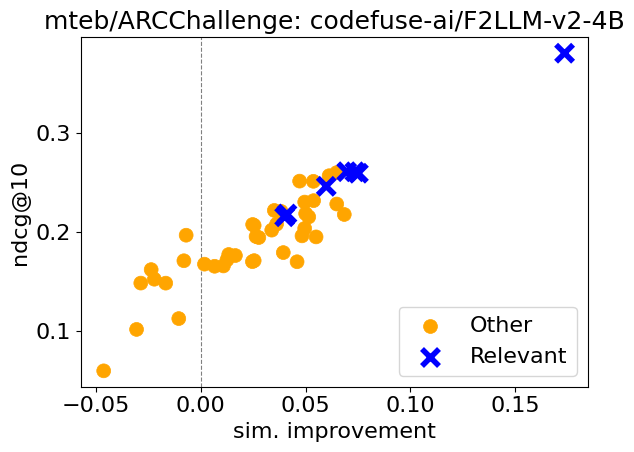}
      \includegraphics[width=0.33\linewidth]{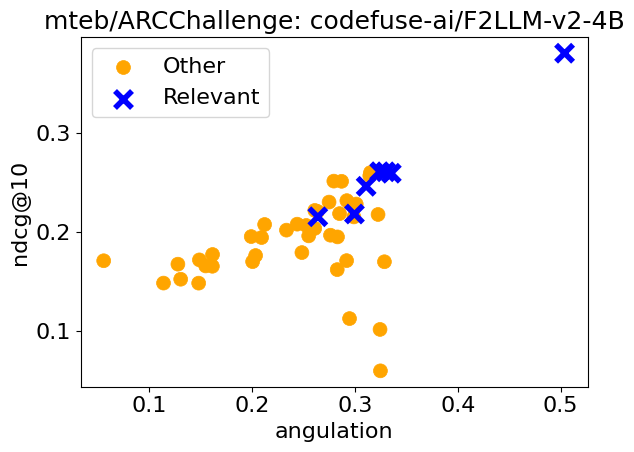}
    \caption{Displacement, similarity improvement, and angulation on outlier codefuse-ai/F2LLM-v2-4B against ndcg@10. Best performing and in all metrics clearly distinguishable instruction is ``Retrieve the answer to the question.''}
    \label{fig:outlier-codefuse}
\end{figure}

\begin{figure}
    \centering
    \includegraphics[width=0.6\linewidth]{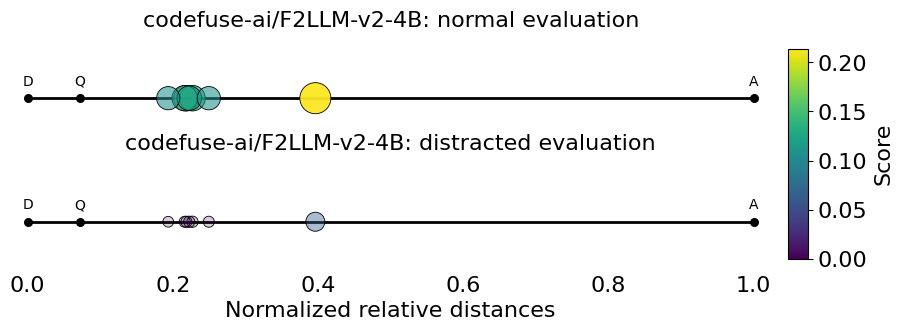}
    \caption{Relative distance plot for codefuse-ai/F2LLM-v2-4B on recall@1: While the best-performing instruction displays large displacement, high similarity increase, and angulation toward the correct target, distractors still inhibit performance greatly. }
    \label{fig:outlier-codefuse-lineplot}
\end{figure}

Figure \ref{fig:outlier-codefuse} presents the results for the outlier model, codefuse-ai/F2LLM--v2-4B, on ndgc@10. Especially for similarity improvement and angulation, high values indicate high performance on relevant instructions. The relevant instruction with the highest displacement, similarity improvement, and angulation is by far the best performing. Despite this, when evaluating with the distractors and recall@1, Figure \ref{fig:outlier-codefuse-lineplot} shows that even the best instruction cannot deal with distractor evaluation fully.

In Table \ref{tab:most_improved_instructions} we present the instructions that most improved their performance in distractor finetuning. Naturally, the finetuning instructions are consistently the instructions with the most improvement, but the improvements are also substantial for other relevant instructions. Furthermore, all but one of the instructions listed are relevant to the task, indicating that fine-tuning was successful in incentivizing the model to react to instructions. 

\begin{table}[]
\centering
\caption{Top 3 instructions that improved their performance (Recall$@$1) the most during fine-tuning.}
\label{tab:most_improved_instructions}
\begin{tabular}{@{}lll@{}}
\toprule
Dataset                & Most improved instructions                                           & Impr. \\ \midrule
\multirow{3}{*}{Tatoeba:cmn} & Retrieve the corresponding translation in Mandarin Chinese.          & 0.49       \\
                         & Given an English sentence, find its translation in Mandarin Chinese. & 0.38       \\
                         & Retrieve the corresponding translation.                              & 0.30       \\ \midrule
\multirow{3}{*}{Tatoeba:fin} & Retrieve the corresponding translation in Finnish.                   & 0.23        \\
                         & Given an English sentence, find its translation in Finnish.          & 0.21       \\
                         & Translate to Finnish.                                                & 0.16       \\ \midrule
\multirow{3}{*}{Tatoeba:vie} & Retrieve the corresponding translation in Vietnamese.                & 0.46       \\
                         & Retrieve the corresponding translation.                              & 0.35       \\
                         & Given an English sentence, find its translation in Vietnamese.       & 0.30       \\ \midrule
\multirow{3}{*}{Tatoeba:ara} & Retrieve the corresponding translation in Arabic.                    & 0.35       \\
                         & Given an English sentence, find its translation in Arabic.           & 0.25       \\
                         & Retrieve parellel sentences in Arabic.                               & 0.21        \\ \midrule
\multirow{3}{*}{Tatoeba:tur} & Retrieve the corresponding translation in Turkish.                   & 0.24       \\
                         & Given an English sentence, find its translation in Turkish.          & 0.21       \\
                         & Translate to Turkish.                                                & 0.19       \\ \midrule
\multirow{3}{*}{Tateoba:fra} & Retrieve the corresponding translation in French.                    & 0.53       \\
                         & Translate to French.                                                 & 0.40       \\
                         & Given an English sentence, find its translation in French.           & 0.40       \\ \midrule
\multirow{3}{*}{Tatoeba:spa} & Retrieve the corresponding translation in Spanish.                   & 0.35       \\
                         & Retrieve the corresponding translation.                              & 0.32       \\
                         & Given an English sentence, find its translation in Spanish.          & 0.22       \\ \midrule
\multirow{3}{*}{Tatoeba:deu} & Retrieve the corresponding translation in German.                    & 0.41       \\
                         & Retrieve the corresponding translation.                              & 0.35       \\
                         & Given an English sentence, find its translation in German.           & 0.23       \\ \midrule
\multirow{3}{*}{SQuAD}   &  Given a question, retrieve Wikipedia passages that answer the question.  &  0.46 \\
&  Given a question, find a related document.  &  0.39 \\
    &  Find documents related to the topic.  &  0.33 \\ \bottomrule
\end{tabular}
\end{table}

\begin{table}[]
    \centering
    \small
    \caption{Fine-tuned performance for ARCChallenge: The baseline performance for ARCChallenge is low, but recall performance with paraphrase distractors increases while rank-based NDCG goes down. Note that the rank-based approach does not reveal distractor performance as strongly as recall@1.}
    \label{tab:after_finetuning_arcchallenge}
\begin{tabular}{@{}lllllll@{}}
    \toprule
                    & \multicolumn{2}{c}{Recall$@$1} & \multicolumn{2}{c}{Recall$@$5} & \multicolumn{2}{c}{NDCG$@$10} \\ \cmidrule(r){2-3} \cmidrule(l){4-5}
\cmidrule(l){6-7}    
Model           & Normal   & Paraphr.            & Normal   & Paraphr.            & Normal & Paraphr.             \\\midrule
    baseline        & 0.059    & 0.000               & 0.183    & 0.138               & 0.139  & 0.097                \\
    SQuAD-finetuned & 0.051    & 0.003($\uparrow$)   & 0.173    & 0.142($\uparrow$)   & 0.130  & 0.094 ($\downarrow$) \\ \bottomrule
    \end{tabular}
\end{table}

\section{Fine-Tuning Details}
\label{app:fine-tuning-details}

\subsection{SQuAD Training Data}

We use SQuAD v1.1 with the provided train / eval split, reserving the evaluation set for final results only. The original training set is divided into three subsets: 10,572 examples for experiments with random-question distractors, matching the number of paraphrase distractors, 2,009 examples for development and checkpoint selection, and the remaining 75,018 examples for fine-tuning.

Each examples produces two fine-tuning instances, one for QA and one for STS, resulting in 150,036 training examples. Because multiple SQuAD questions can share the same answer paragraph, negatives are sampled only from examples with different answer paragraphs to avoid introducing false negatives.

\subsection{Tatoeba Training Data}

Since MTEB contains only test data, we obtain training data from OPUS Tatoeba collection (v2026-07-08) \citep{tiedemann-2012-parallel}. We first deduplicate the data so that each source and target sentence appears only once. To avoid accidental train--test overlap, we remove pairs whose source or translation appears in any language's test set. We also remove pairs for which either sentence has a similarity above 0.8 to any test sentence in any language. Similarity is measured as the cosine similarity between TF-IDF vectors based on character 4-grams, after lowercasing and removing pucntuation. After filtering, we randomly sample up to 10,000 pairs per language, using 1,000 for development and the remaining for training. Vietnamese has only 7,913 pairs after filtering, resulting in 6,913 training and 1,000 development examples. All other languages contain 9,000 training and 1,000 development examples. All test sets contain 1,000 examples. For the random-distractor experiment, we additionally sample 1,000 English sentences from monolingual English Tatoeba, applying the same filtering against all training, development, and test sets.

\subsection{Fine-Tuning Parameters}

Fine-tuning is performed with the SWIFT framework \citep{zhao2024swiftascalablelightweightinfrastructure} using full-parameter fine-tuning, an InfoNCE loss, and cosine learning rate scheduler. We fine-tune for one epoch and save a checkpoint every 1,000 steps. The final checkpoint is selected based on performance in the paraphrase distractor setting. We use per-device training batch size is of 8, gradient accumulation of 1, a learning rate of 1e-6, a warmup ratio of 0.05, weight decay of 0.01, InfoNCE temperature of 0.05, and a maximum sequence length of 1,024 tokens.

We do not use fake-negative masking, a similarity-margin filtering technique commonly used to avoid false negatives, because this could mask the query-side distractors that we explicitly want the model to learn from. Query-side negatives are included in 50\% of the training examples to balance learning from these hard negatives while preserving standard retrieval behavior. Because query-side negatives are assumed to have considerably higher similarity, they can contribute disproportionately to the InfoNCE objective, where softmax weighting emphasizes high-similarity negatives.

\subsection{Fine-Tuning Instructions}

The following instructions are used in fine-tuning:
\begin{itemize}
    \item SQuAD QA: \emph{Given a question, retrieve Wikipedia passages that answer the question.}
    \item SQuAD STS: \emph{Given a question, retrieve questions that are semantically equivalent to the given question.}
    \item Tatoeba bitext: \emph{Retrieve the corresponding translation in \{lang\}}.
    \item Tatoeba STS: \emph{Retrieve semantically similar sentences.}
\end{itemize}

The same task-specific instractions are also used in the main evaluation results.

\section{Language-Specific Results for Tatoeba}
\label{app:tatoeba-results}

\begin{figure}
    \centering
    \includegraphics[width=0.8\linewidth]{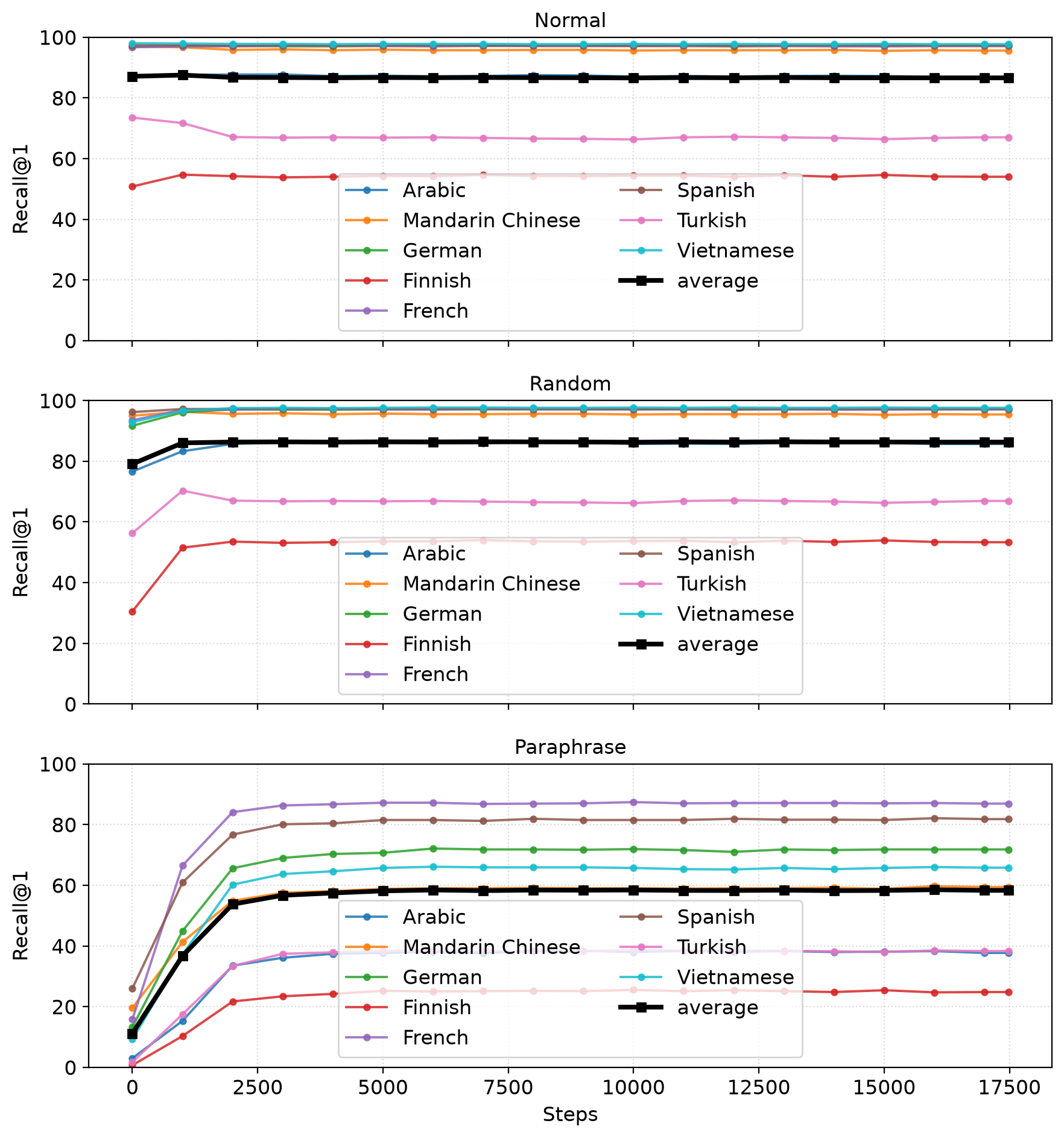}
    \caption{Language-specific results for Tatoeba fine-tuning checkpoints on the devevelopment data.}
    \label{fig:tatoeba-language-results}
\end{figure}

Tatoeba fine-tuning performance across checkpoints is shown separately for each language in Figure \ref{fig:tatoeba-language-results}, together with the average over languages.

\section{Fine-Tuning Without Query-Side Negatives}
\label{app:finetuning-wo-hardnegatives}

\begin{figure}
    \centering
     \begin{minipage}[t]{0.49\linewidth}
     \vspace{0pt}
    \includegraphics[width=\linewidth]{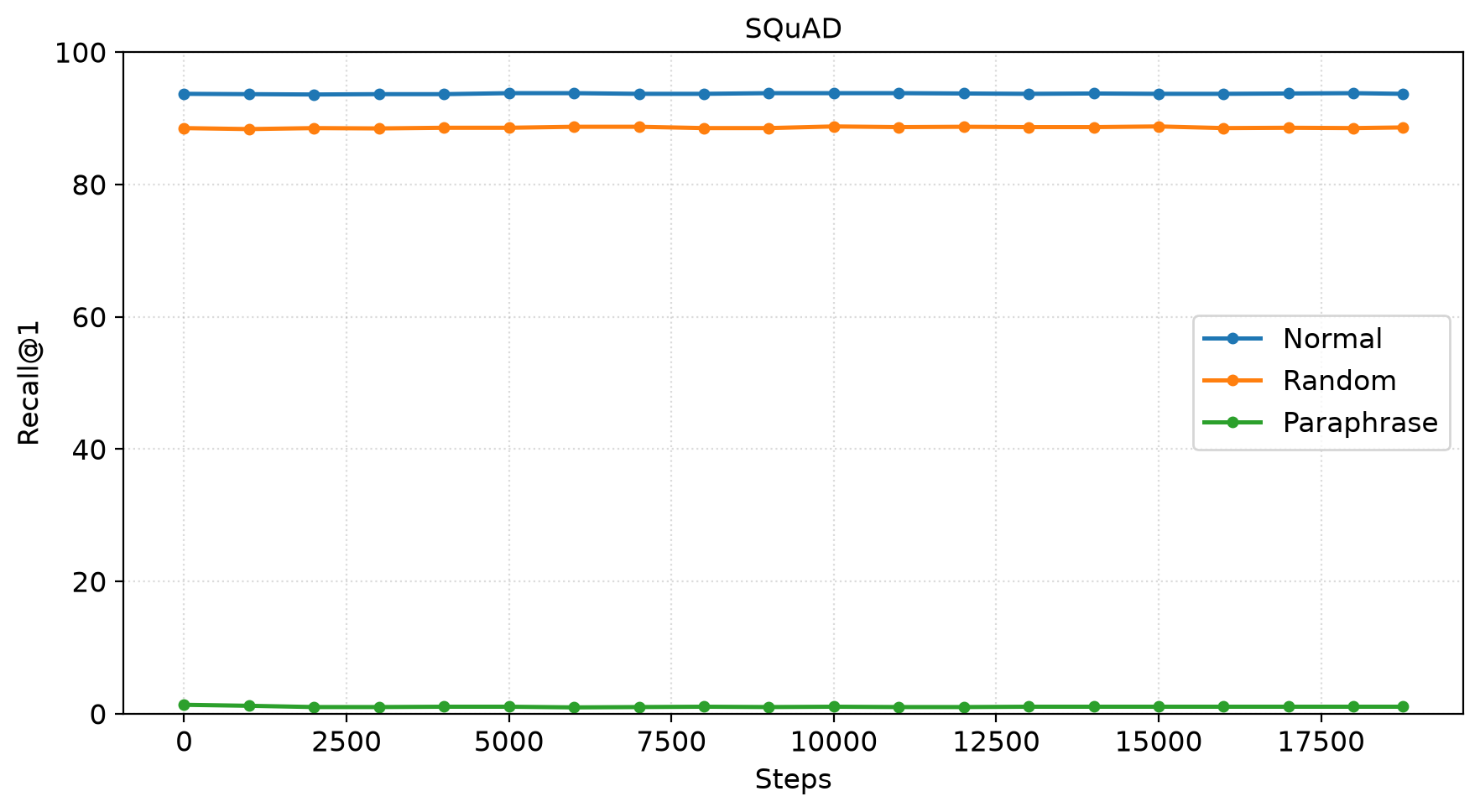}
    \end{minipage}
    \hfill
    \begin{minipage}[t]{0.49\linewidth}
    \vspace{0pt}
   \includegraphics[width=\linewidth]{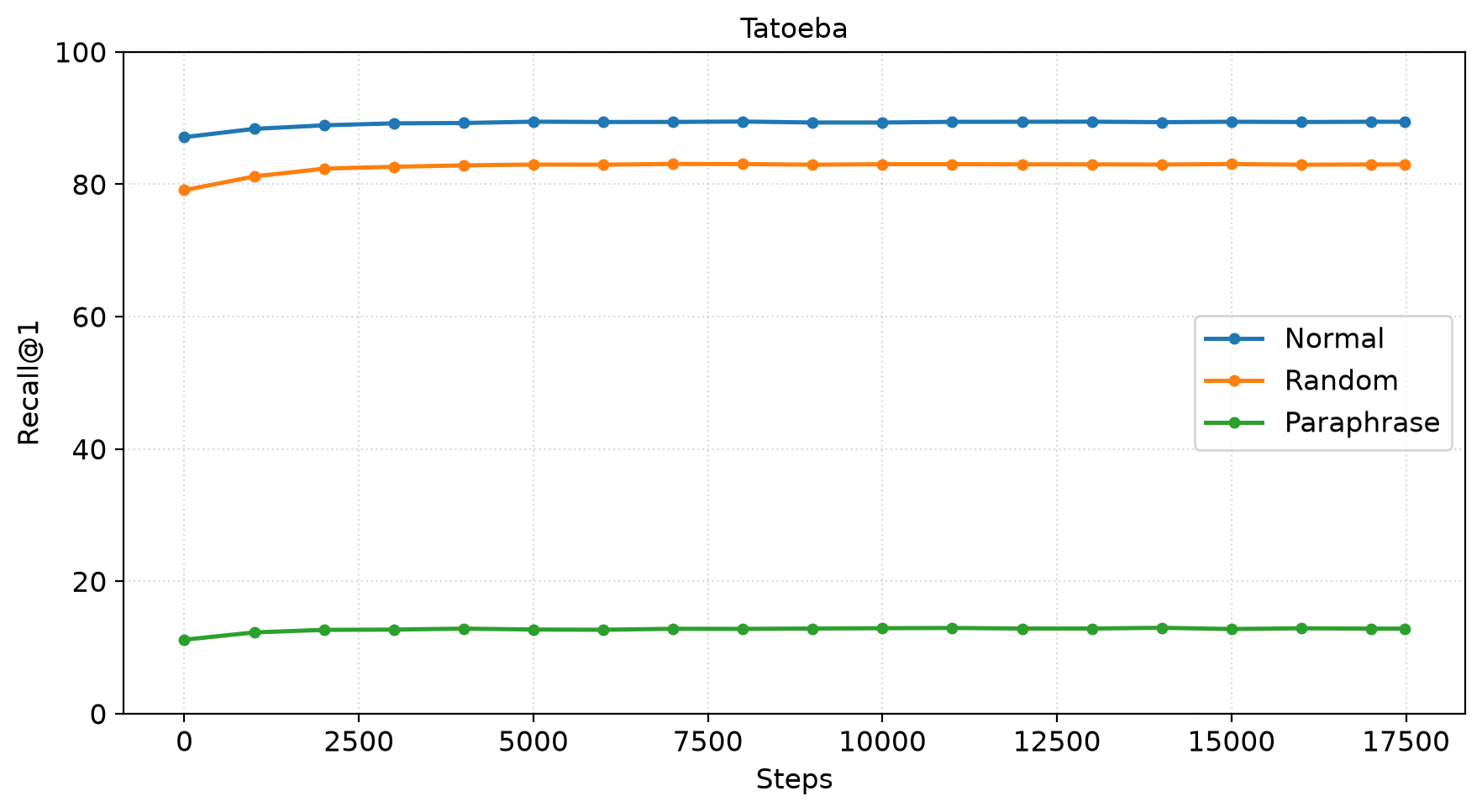}
    \end{minipage}
    \captionof{figure}{Development set performance during SQuAD and Tatoeba fine-tuning without query-side negatives. Continued fine-tuning without inserted distractors does not improve performance in the paraphrase distractor setting.}
    \label{fig:finetuning-wo-hardnegatives}
\end{figure}

To ensure that the improved performance in the paraphrase distractor setting does not result merely from additional task-specific fine-tuning, we also fine-tuned models on the same SQuAD and Tatoeba data using the same training setup, but without query-side negatives. Figure~\ref{fig:finetuning-wo-hardnegatives} shows performance across fine-tuning checkpoints. Models trained without query-side negatives show no improvement in the distractor setting, indicating that the gains result from the added negatives rather than task-specific fine-tuning alone.

\section{Full MTEB results for fine-tuned models}
\label{app:full-mteb-results}

\begin{figure}
    \centering
    \includegraphics[width=0.8\linewidth]{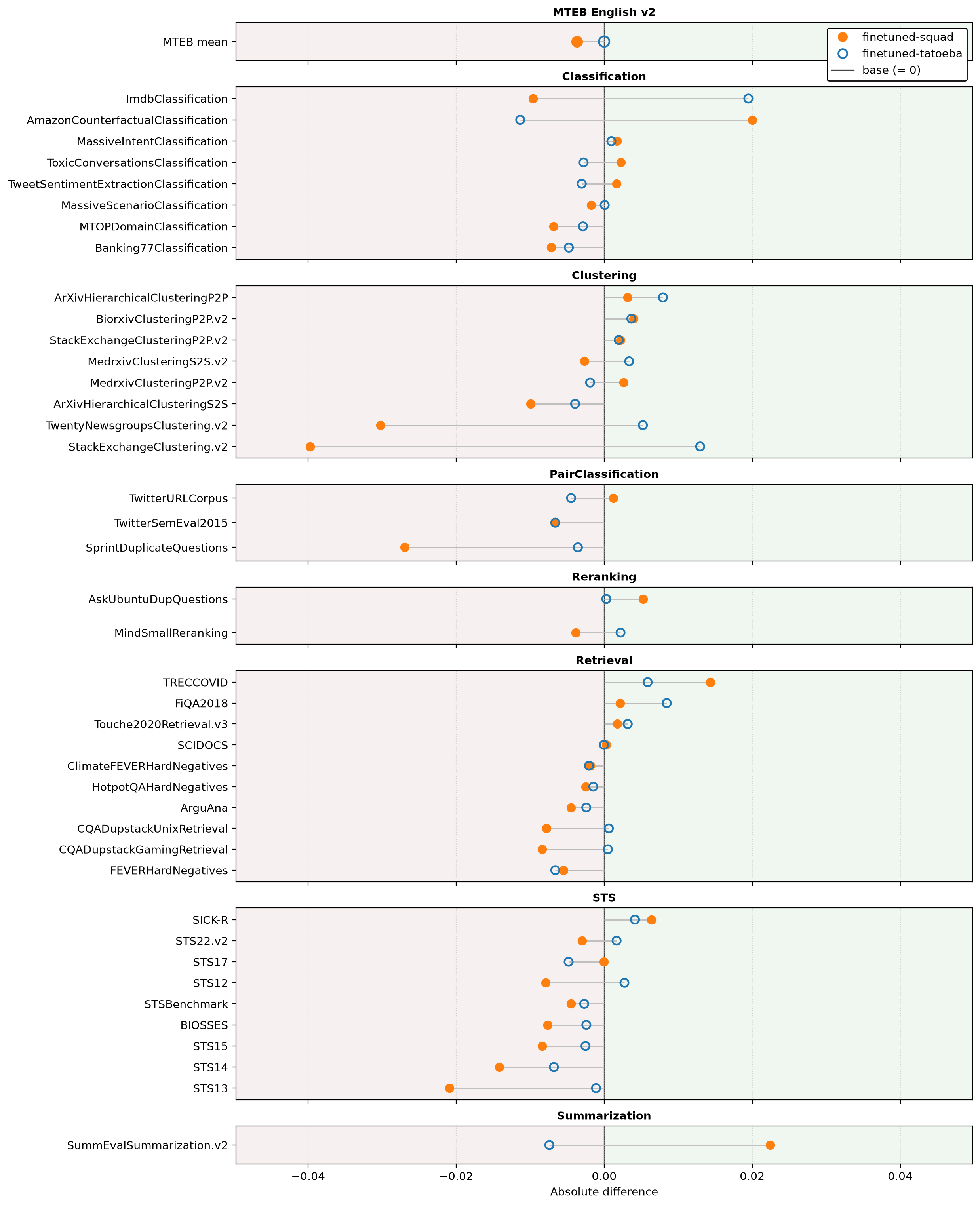}
    \caption{MTEB English v2 benchmark results for the fine-tuned models, compared with the base model. We report the absolute difference, where positive values indicate better performance after fine-tuning.}
    \label{fig:mteb-full-results}
\end{figure}

The full results on the MTEB English v2 benchmark are shown in Figure \ref{fig:mteb-full-results}. The benchmark includes 41 tasks across 7 categories. We report the absolute score difference (fine-tuned score minus base score), with all MTEB scores running between [0,1]. Positive values indicate better performance after fine-tuning. The MTEB mean is the arithmetic mean across all tasks.

\section{Computational Details}
\label{app:computational-details}

We use AMD MI250x GPUs \& Nvidia GH200 Grace Hopper GPUs, and AMD EPYC 7763 \& AMD Turin 9965 CPUs on LUMI and Roihu clusters of CSC -- IT Center for Science, Finland, to run all experiments. 
All the following compute estimates are top-end approximations.

\textit{Fine-tuning} We fine-tune 4 models (two datasets: SQuAD and Tatoeba, two settings: paraphrase distractors and control). Training each model took approximately 12 hours on one GPU.

\textit{Structure} We calculate structural metrics (all instructions together), 8 models and 9 datasets (ARCChallenge + 8 Tatoeba datasets) and additionally we run the fine-tuning comparisons, which include the 2 fine-tuned models, one on SQuAD, one on all Tatoeba languages, and the baseline model for SQuAD: all together 82 combinations. Each combination was run on 2 GPUs, and took approximately 20 to 60 minutes, depending on model size.

\textit{Evaluation} We evaluated the same 82 combinations as above in both standard and distracted setting. We again used 2 GPUs and the evaluation took approximately 45 to 90 minutes for ARCChallenge and Tatoeba (both with \~1000 examples), and approximately 6-7 hours for the two SQuAD (\~10k examples) settings. Running the additional MTEB English v2 benchmark on one GPU took 8 hours per model. 

Altogether, we used at most 504 GPU hours.